\documentclass[review,11pt,3p,number,sort&compress]{elsarticle}
\usepackage{amsmath,amssymb}
 \usepackage[utf8]{inputenc}
 \usepackage[T1]{fontenc}
\usepackage {graphicx} 
\usepackage {float}
\usepackage {booktabs} 
\usepackage {siunitx} 
\usepackage {caption} 
\usepackage{algorithm}
\usepackage{algpseudocode} 

\begin{document}

\begin{frontmatter}
\title{PEFTT: Parameter-Efficient Fine-Tuning for low-resource Tibetan pre-trained language models}

\author[a,b]{ZHOU Mingjun}
    \ead{zhoumj77@outlook.com}
\author[a,b]{DAIQING Zhuoma}
\author[a,b,c]{QUN Nuo\cormark[*]}
    \ead{q\_nuo@utibet.edu.cn}
\author[a,b,c]{NYIMA Tashi}

\affiliation[a]{organization={School of Information Science and Technology, Tibet University},
            addressline={}, 
            city={Lhasa},
            postcode={850000}, 
            state={Tibet},
            country={China}}
\affiliation[b]{organization={Collaborative Innovation Center for Tibet Informatization by MOE and Tibet Autonomous Region},
            addressline={}, 
            city={Lhasa},
            postcode={850000}, 
            state={Tibet},
            country={China}}
\affiliation[c]{organization={Engineering Research Center of Tibetan Information Technology, Ministry of Education, Tibet University},
            addressline={}, 
            city={Lhasa},
            postcode={850000}, 
            state={Tibet },
            country={China}}
\cortext[*]{Corresponding author}
\begin{abstract}In this era of large language models (LLMs), the traditional training of models has become increasingly unimaginable for regular users and institutions. The exploration of efficient fine-tuning for high-resource languages on these models is an undeniable trend that is gradually gaining popularity. However, there has been very little exploration for various low-resource languages, such as Tibetan. Research in Tibetan NLP is inherently scarce and limited. While there is currently no existing large language model for Tibetan due to its low-resource nature, that day will undoubtedly arrive. Therefore, research on efficient fine-tuning for low-resource language models like Tibetan is highly necessary. Our research can serve as a reference to fill this crucial gap. Efficient fine-tuning strategies for pre-trained language models (PLMs) in Tibetan have seen minimal exploration. We conducted three types of efficient fine-tuning experiments on the publicly available TNCC-title dataset: "prompt-tuning," "Adapter lightweight fine-tuning," and "prompt-tuning + Adapter fine-tuning." The experimental results demonstrate significant improvements using these methods, providing valuable insights for advancing Tibetan language applications in the context of pre-trained models.
\end{abstract}
\begin{keyword}
Tibetan\sep PLM\sep Prompt-tuning\sep Adapter
\end{keyword}
\end{frontmatter}
    
    \section{Introduction}
    The technique of full fine-tuning is currently the mainstream approach when adapting PLMs to various downstream language tasks. It involves updating and storing the parameters of the entire model. However, with the increasing size of models, using full fine-tuning may become excessively costly. The introduction of Bert \cite{devlin-bert-2019} enabled NLP to perform adaptive fine-tuning for downstream tasks, similar to the paradigm in computer vision (CV), leading the way for fine-tuning in NLP downstream task training. However, even the parameter scale of the Bert-base model has reached hundreds of millions. Subsequently, the GPT2 \cite{radford-language-nodate} model increased its parameter count to 1.5 billion, while the 11B version of the T5 \cite{raffel-exploring-2020} model reached billions, making effective training challenging on consumer-grade hardware. The GPT3 \cite{brown-language-nodate} model has a staggering 175 billion training parameters, making it almost unimaginable for regular individuals or institutions to train such large models. For the current experiment, fully fine-tuning the cino-large-v2 model \cite{yang-cino-2022} (with 440 million parameters) was restricted by insufficient GPU memory, and the cino-base-v2 model (with 190 million parameters) required a smaller batch size to handle long-text tasks. This hardware limitation is not only pertinent to general users; even large companies like OpenAI face obstacles in model iteration and upgrading due to these constraints. Therefore, research on lightweight fine-tuning for large-scale models and significant performance improvement under relatively small parameter increases holds great significance for pre-trained models across various domains. Adapter and Prompt-tuning are two efficient fine-tuning methods that can meet the aforementioned requirements. This paper explores these two methods on Tibetan PLMs. In the Prompt-tuning experiments, significant performance improvements were achieved. In the Adapter experiments, lightweight fine-tuning resulted in relatively small average performance loss. Finally, in the experiment that combined Prompt-tuning and Adapter, the combined approach achieved the effect of lightweight fine-tuning while improving evaluation metrics. Our research makes the following contributions and references to Tibetan NLP and similar low-resource language NLP:
\begin{itemize}
    \item We investigate three efficient fine-tuning approaches for Tibetan pre-trained models and achieve highly promising results.
    \item Various efficient fine-tuning methods are popular in high-resource languages like English, such as prompt-tuning. Although tools like OpenPrompt make it convenient to construct prompt-tuning training, their support for Tibetan and even Chinese is severely lacking. We adapt the toolkit to our local environment and corresponding models. Specifically, for training models in the Tibetan monolingual context, we dynamically expand the Tibetan vocabulary during training and inference.
    \item Considering the differences in various pre-trained models, we conduct numerous configuration experiments. For each model, we explore efficient fine-tuning configurations that enhance model adaptability. These configurations can serve as references for efficient fine-tuning in various low-resource language NLP applications, not limited to Tibetan.
\end{itemize} 
    \section{Related work}
    This section elaborates on the research and development of Adapter technology, the research and development of Prompt-tuning technology, and the exploration of these two techniques in the field of Tibetan NLP.
    \subsection{Adapter}
    Neil Houlsby et al. \cite{houlsby-parameter-efficient-2019}discovered the inefficiency of fine-tuning large models and introduced the Adapter module for the first time in their research. This module adds only about 3\% additional parameters and freezes most of the model's parameters, yet achieves performance comparable to full fine-tuning. This finding not only garnered widespread attention in the NLP field but also sparked research into lightweight fine-tuning operations using Adapter modules in the computer vision (CV) domain. For instance, Taojiannan Yang et al. \cite{yang-aim-2023} identified the high cost and unnecessary nature of fine-tuning video models in the CV domain. They introduced Adapter modules to existing relevant models and achieved better spatiotemporal understanding by introducing flexible temporal, spatial, and spatiotemporal adaptive adjustments. This improvement enhanced the model's performance with minimal fine-tuning of parameters. In the NLP field, Xiang Lisa Li et al. \cite{li-prefix-tuning-2021} proposed the prefix-tuning method, which freezes all model parameters and fine-tunes only 0.1\% of the parameters relative to the original model by optimizing task-specific prefix vectors, achieving competitive performance. Edwardhu et al. \cite{hu-lora-2021} introduced LoRA, , an adaptive Adapter module with low-rank characteristics, incorporating trainable rank decomposition matrices into every layer of the Transformer. It achieved excellent performance while significantly reducing the number of parameters and GPU memory requirements, even outperforming full fine-tuning on multiple models. Haokun Liu et al. \cite{liu-few-shot-2022} found that Adapter technology can address the computational and memory overhead challenges in context learning on small samples. They proposed the IA$^3$ Adapter module structure, achieving the best results in the RAFT test \cite{alex-raft-2022} at the time. Jonas Pfeiffer et al. \cite{pfeiffer-mad-x-2020} introduced MAD-X, an Adapter architecture addressing the challenge of limited transferability of multilingual BERT to low-resource languages. These research efforts and applications are of significant importance in advancing and applying Adapter technology in various domains.
    \subsection{Prompt-tuning}
    For an extended period, the prevailing approach has been fine-tuning pre-trained models for various downstream tasks, showcasing remarkable performance. This strategy has stood as the mainstream paradigm in the NLP field. However, the downstream task fine-tuning objectives may differ significantly from the model's pre-training objectives, limiting the model's knowledge exploitation capabilities \cite{liu-gpt-2021}. To address this, the authors of GPT-3, including Brown et al., introduced a new fine-tuning paradigm called Prompt-tuning. This method uses natural language and task examples as context, making downstream tasks resemble language modeling. Subsequently, various studies related to Prompt-tuning emerged. Schick et al. conducted research on artificial Prompt templates (also known as hard templates) in 2020 and 2021 \cite{schick-automatically-2020,schick-exploiting-2021}, achieving good results in text classification and reasoning tasks. However, designing appropriate hard templates requires rich domain knowledge, leading researchers to explore generative hard template methods. For instance, Shin et al. \cite{shin-autoprompt-2020} developed a task-adaptive template auto-generation method, while Tianyu Gao et al. \cite{gao-making-2021} designed an automatic prompt generation method that performed well in low-resource environments. Ben-David et al. \cite{ben-david-pada-2022} introduced PADA based on the T5 model, a method for online adaptive prompt generation across any domain. The objective behind template design is to enhance the performance of pre-trained language models in targeted downstream tasks. As research progressed, a superior approach in certain cases emerged, involving soft prompt templates, where Prompt-tuning is achieved by setting multiple learnable vectors. In this case, the research scope expanded to encompass non-human-understandable natural language. Lester et al. and Vu et al. \cite{lester-power-2021,vu-spot-2022} were among the earliest researchers in this field. In addition to the mentioned studies, other research on Prompt-tuning includes methods such as prefix-tuning combined with Adapters, as well as P-tuning proposed by Xiao Liu et al. \cite{liu-p-tuning-2022}.
    \subsection{Research on Prompt-tuning and Adapter in Tibetan NLP}
    In high-resource language domains, such as English, Prompt-tuning and Adapter technologies have undergone years of development and achieved significant results. However, in the field of NLP for low-resource languages like Tibetan, research and application of these two efficient fine-tuning techniques are nearly nonexistent. Hence, a pressing necessity exists for research endeavors that can fill the existing gap in the domain of natural language processing, especially concerning low-resource languages.
    
    \section{Models\&Methodology}
    In this study, we utilize Tibetan-specific pre-trained language models (PLMs) with well-defined sources: CINO, Tibert, and Tibetan-bert.\cite{liu-tibert-2022,zhang-research-2022} We employ three efficient fine-tuning techniques: Prompt-tuning, Adapter, and the combination of Prompt-tuning and Adapter, to conduct experiments on these three PLMs.
    \subsection{Models}
    \subsubsection{CINO}
    CINO, currently in its second version, is based on XLMRoberta\cite{conneau-unsupervised-2020}. CINO supports eight Chinese languages, including Tibetan, and has a significantly larger vocabulary of 135,359 tokens. It offers three versions: small, base, and large, with transformer layers\cite{vaswani-attention-2017} of 6, 12, and 24, respectively. Among Tibetan PLMs, it achieves the best performance on the TNCC\cite{qun-end--end-nodate} document-tibetan.txt dataset (referred to as TNCC-d). 
    \subsubsection{Tibert}
    Tibert is a pre-trained language model developed by the NLP research team at Minzu University of China. Utilizing the BERT-base architecture as its foundation, the model is constructed with 12 Transformer layers and employs a vocabulary size of 30,005,Tibert covers 99.5\% of Tibetan vocabulary. In downstream task evaluations on the TNCC-t and TNCC-d Tibetan datasets, Tibert demonstrates highly competitive performance. It achieves macro-F1 evaluation scores of 61.72\% and 70.94\% on the test sets, respectively.
    \subsubsection{Tibetan-bert}
    Tibetan-bert, developed by the UTibetNLP team at Tibet University, is based on the BERT-base architecture similar to Tibert. It also comprises 12 Transformer layers and has a vocabulary size of 32,267.
    \subsection{Methodology}
    \subsubsection{Prompt-tuning}
    \begin{figure}[ht!]
          \centering
          \includegraphics[width=1\linewidth]{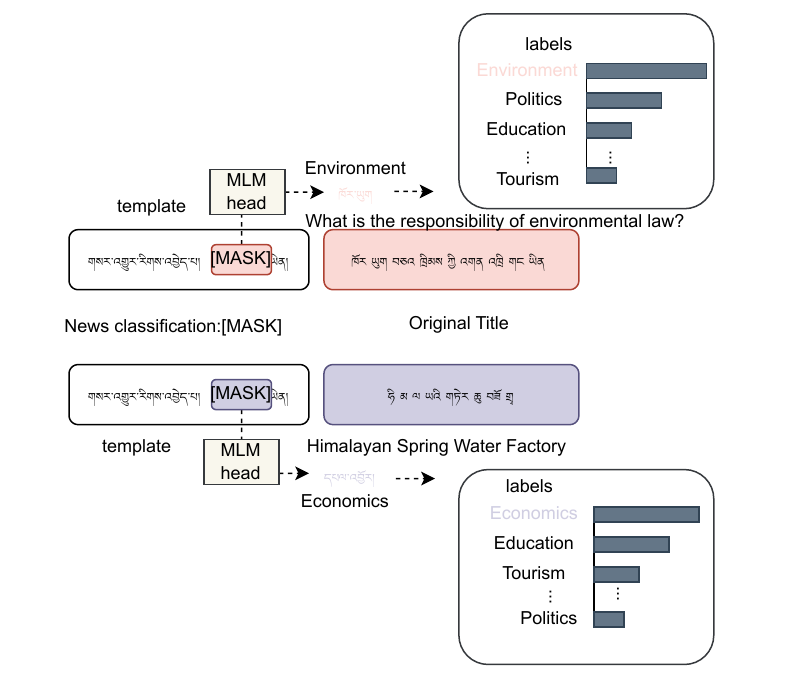} 
          \caption{\label{fig:prompt} Prompt-tuning using manual template} 
    \end{figure}
    For the Prompt-tuning technique, we employ the open-source framework called OpenPrompt\cite{ding-openprompt-2022} developed by Tsinghua University. This framework provides various tools and libraries for Prompt-tuning, allowing users to conveniently construct training frameworks that integrate Prompt-tuning components and PLMs. Users can customize and select templates and verbalizers. In this study, we leverage this platform and perform efficient fine-tuning using manual templates (hard templates). The composition and principles of this approach are illustrated in Figure \ref{fig:prompt} Prompt-tuning is a method that adds additional information to a pre-trained language model (PLM) by reformulating downstream tasks as cloze-style questions\cite{taylor-cloze-1953}. This methodology incorporates templates and a set of labeled words. The template offers contextual background for the current task, while the labeled words represent high-probability vocabulary predicted by the PLM based on the context. In the case of the TNCC-t twelve-classification dataset for this experiment, an input sentence $x$ is represented as: $$x = [w_1, w_2, w_3, ..., w_{n-1}, w_n]$$, and the output label is $y$, where $$y \in [y_0, y_1, y_2, ..., y_{n-1},y_n]$$, and the label space $$Y = \{Politics, Education, ..., Tourism\}$$, encompassing twelve labels.
    Prompt-tuning formalizes the classification task as a Masked Language Model (MLM) task. Given a PLM $\large M$ and a vocabulary set $V$, the input sentence $x$ is transformed into a prompt input $x_{prompt} = T(x)$ using a template function $T(.)$ with a <MASK> token. Additionally, a set of label words $V^* \subset V$ constitutes the prompt, along with a verbalizer function v that maps each label to the vocabulary $V: L \rightarrow V$.
    We can transform the current classification task into addressing the following: Given an input $x$, we obtain $T(x)$. Then, employing $\large M$, we ascertain the label $y \in L$ such that $v(y)$ is the most probable replacement for <MASK>. Concretely, $$x = (a, b); T(a, b) = (a <mask>, b)$$ where 'a' represents our template, and 'b' represents the original text. Combined with the verbalizer function $v$, it maps each label with meaning $[y_0, y_1, y_2, ..., y_{n-1}]$to the label space $$Y = \{Politics, Education, ..., Tourism\}$$. 
    For instance, given the input: 
    \begin{center}
    \includegraphics[height=1.5em]{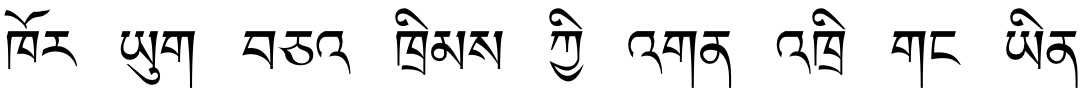}
    \end{center} 
    Then,transform the input $T(x)$ with template text and a mask.
    \begin{center}
    $T(x)$ = \includegraphics[height=1.25em]{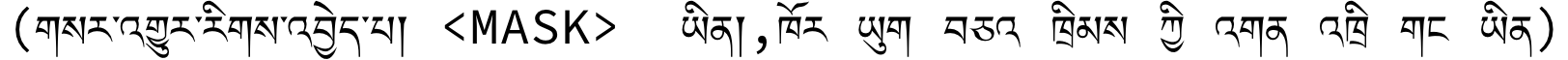}
    \end{center} 
    In this case, the task is no longer to assign fixed-meaning labels, but rather to answer in the context of the entire sentence. For example:
    \begin{center}
    \includegraphics[height=1.5em]{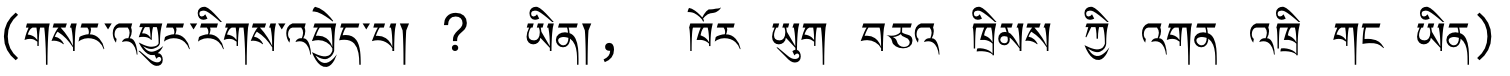}
    \end{center}
    Ultimately, the most likely Environment is selected from $ Y = \{Politics, Education, ..., Tourism\}$. Translating this sentence combined with the template to English:
    \begin{center}
    News Classification:[MASK]. What is the responsibility of environmental law?
    \end{center} 
    \subsubsection{Adapter}
    \begin{figure}[ht!]
          \centering
          \includegraphics[width=1\linewidth]{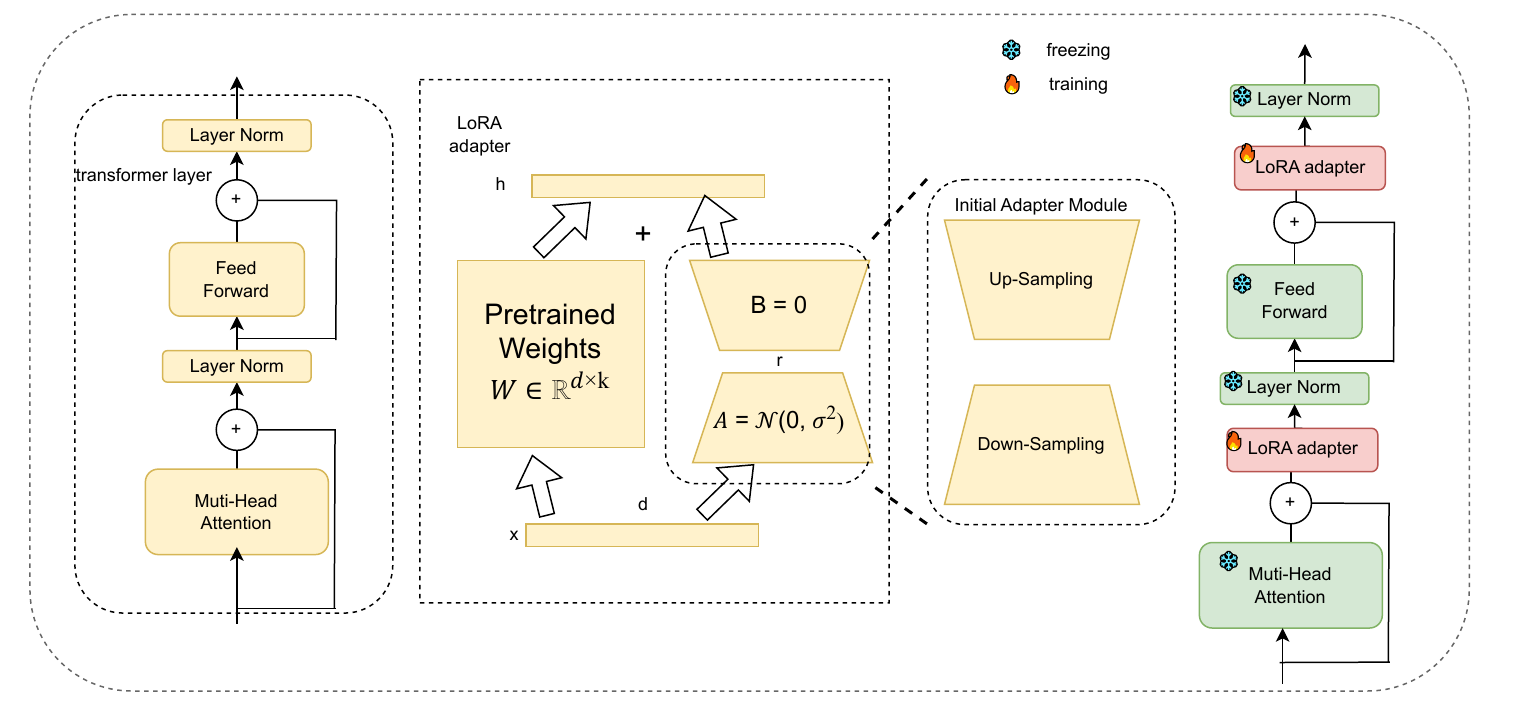} 
          \caption{\label{fig:adpter} Customized Composition of Adapter} 
    \end{figure}
    For the Adapter support framework, we have chosen the open-source framework OpenDelta \footnote{https://github.com/thunlp/OpenDelta}developed by Tsinghua University. This framework provides customizable options for Adapter types, structures, and injection positions, and it integrates well with the OpenPrompt framework. We have opted for the LoRA (Low-Rank Adaptation) type of Adapter and injected it after the Self Attention output and after the FeedForward output, as depicted in Figure \ref{fig:adpter}.
    For each received lower-layer output x, the LoRA module processes as follows: Given that the training weights $$W_0 \in R^{d\times k}$$ exhibit low-rank characteristics\cite{aghajanyan-intrinsic-2021}, specifically $r \ll min(d, k)$, we can represent the update by performing a low-rank decomposition as $$W_0 + \Delta W = W_0 + B\times A$$, where $B \in R^{d\times r} and A \in R^{r\times k}$. Here, $W_0$ remains frozen during training. Consequently, for the input x, the forward pass shifts from$ h = W_0 x $ to $$ h = W_0 x + \Delta Wx = W_0 x + B\times Ax$$
    Compared to the initial Adapter module, LoRA shares some similarities but with a notable distinction. Both LoRA and the original Adapter employ a trapezoidal module with up-sampling and down-sampling, as depicted in Figure \ref{fig:adpter}, where $B \in R^{d\times r} $and $A \in R^{r\times k}$. However, LoRA distinguishes itself by adopting a parallel relationship with the training weights $W_0 \in R^{d\times k}$, while the original Adapter follows a hierarchical relationship by directly freezing the lower-layer$W_0 \in R^{d\times k}$  and sampling it. This results in a two-stage forward propagation: first, $y = W_0 x$, followed by $h = B\times Ay$.
    
    \section{Dataset\&Performance Metrics}
    \subsection{Dataset}
    \begin{figure}[ht!]
          \centering
          \includegraphics[width=0.5\linewidth]{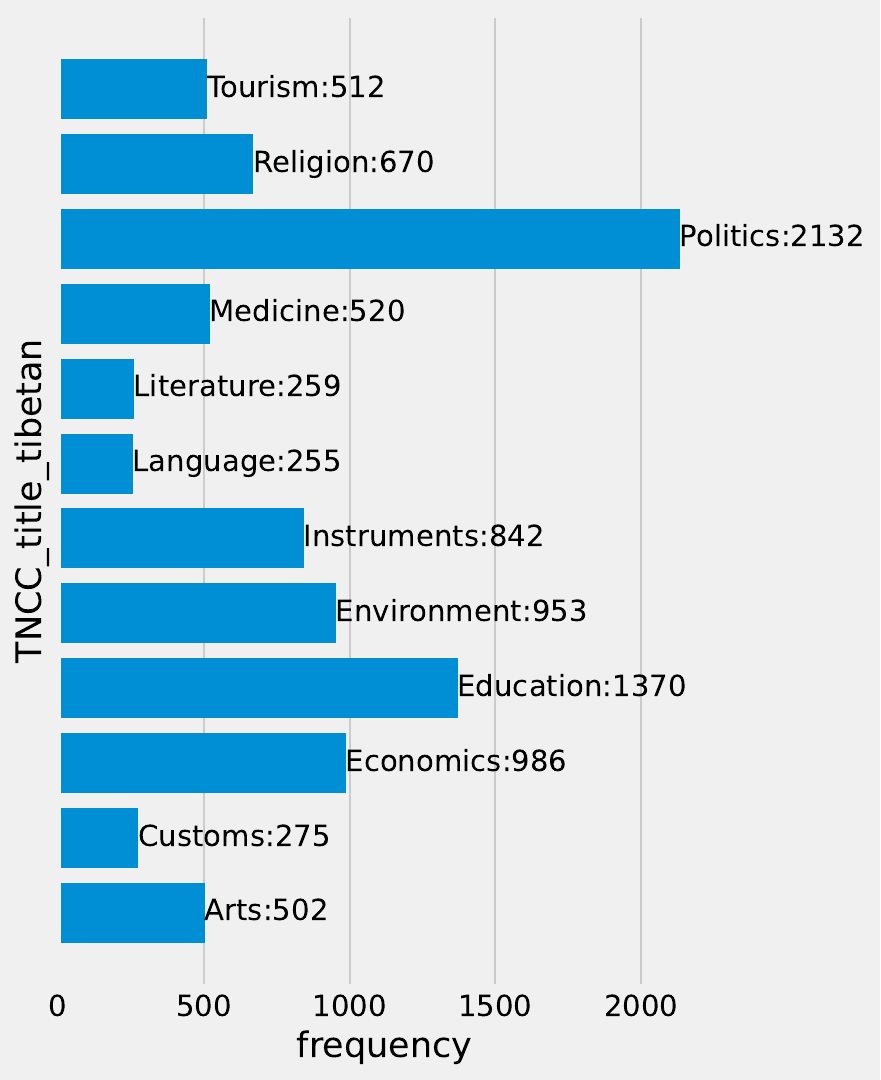} 
          \caption{\label{fig:dataset} The data distribution of TNCC\_title\_tibetan} 
    \end{figure}
    The dataset used in this experiment is the Tibetan News Classification Corpus (TNCC-t), a publicly available dataset. The dataset has been partitioned into training, validation, and test sets, distributed in an 8:1:1 ratio. The distribution of the dataset can be referred to in Figure \ref{fig:dataset}.
    It's worth noting that the TNCC-t dataset exhibits an uneven distribution due to limitations in data sources. The majority of news data is sourced from government-related websites, leading to a relatively higher number of news headlines in the political domain.
    \subsection{Performance Metrics}
    In this experiment, Accuracy and Macro-F1 are selected as performance evaluation metrics. Given the non-uniform distribution of the TNCC-t dataset, Macro-F1 holds significant value in assessing model performance. While Accuracy indicates the ratio of correctly classified samples to the total samples, the Macro-F1 metric takes into account the importance of each class, providing an average F1 score across all classes. This evaluation approach offers a comprehensive measure of the model's performance across different categories. The specific calculation formulas can be referred to in the definitions provided in Table \ref{tab:metric}, where TP denotes true positives, TN denotes true negatives, FP denotes false positives, and FN denotes false negatives.
    \begin{table}[ht!]
          \centering
          \resizebox{.8\linewidth}{!}{
            \begin{tabular}{cc}
            \toprule
            Performance Metrics &  Calculation Formulas \\
            \midrule
           Accuracy & $\text{Accuracy} = \frac{\text{TP} + \text{TN}}{\text{TP} + \text{TN} + \text{FP} + \text{FN}}$ \\
           \midrule
          Macro-f1 & 
          $\begin{aligned}
            \text{precision}_i &= \frac{\text{TP}_i}{\text{TP}_i + \text{FP}_i} \\
            \text{recall}_i &= \frac{\text{TP}_i}{\text{TP}_i + \text{FN}_i} \\
            \text{f1}_i &= \frac{2\text{precision}_i \text{recall}_i}{\text{precision}_i + \text{recall}_i} \\
            \text{Macro-f1} &= \frac{1}{n} \sum_{i=1}^{n} \text{f1}_i
          \end{aligned}$ \\
          \bottomrule
            \end{tabular}%
    }
          \caption{Mathematical Formulation of Performance Evaluation Metrics}\label{tab:metric}%
    \end{table}%
    
    \section{Experiment and Analysis}
    The experiment will be conducted based on the Model Methods section and the Dataset and Performance Metrics section.
    \subsection{Abbreviation Introduction}
    To simplify the expression in subsequent experiments and avoid repeated use of lengthy experiment scenario names, an abbreviation approach is adopted to substitute various experiment scenario names. Specific details can be found in Table \ref{tab:abb} .
    \begin{table}[ht!]
      \centering
      \resizebox{1\linewidth}{!}{
        \begin{tabular}{cc}
          \toprule
          Abbreviation & Meaning \\
          \midrule
          CSW & CINO-small\_wo, CINO-small fully fine-tuning \\
          CSP & CINO-small\_Prompt, CINO-small Prompt-tuning \\
          CSA & CINO-small\_Adapter, CINO-small Adapter lightweight fine-tuning \\
          CSAP & CINO-small\_Adapter\_Prompt, CINO-small Adapter lightweight Prompt-tuning \\
          CBW & CINO-base\_wo, CINO-base fully fine-tuning \\
          CBP & CINO-base\_Prompt, CINO-base Prompt-tuning \\
          CBA & CINO-base\_Adapter, CINO-base Adapter lightweight fine-tuning \\
          CBAP & CINO-base\_Adapter\_Prompt, CINO-base Adapter lightweight Prompt-tuning \\
          CLW & CINO-large\_wo, CINO-large fully fine-tuning \\
          CLP & CINO-large\_Prompt, CINO-large Prompt-tuning \\
          CLA & CINO-large\_Adapter, CINO-large Adapter lightweight fine-tuning \\
          CLAP & CINO-large\_Adapter\_Prompt, CINO-large Adapter lightweight Prompt-tuning \\
          TW & Tibert\_wo, Tibert fully fine-tuning \\
          TP & Tibert\_Prompt, Tibert Prompt-tuning \\
          TA & Tibert\_Adapter, Tibert Adapter lightweight fine-tuning \\
          TAP & Tibert\_Adapter\_Prompt, Tibert Adapter lightweight Prompt-tuning \\
          TBW & Tibetan-bert\_wo, Tibetan-bert fully fine-tuning \\
          TBP & Tibetan-bert\_Prompt, Tibetan-bert Prompt-tuning \\
          TBA & Tibetan-bert\_Adapter, Tibetan-bert Adapter lightweight fine-tuning \\
          TBAP & Tibetan-bert\_Adapter\_Prompt, Tibetan-bert Adapter lightweight Prompt-tuning \\
          PLM\_wo & PLM fully fine-tuning \\
          PLM\_p & PLM Prompt-tuning \\
          PLM\_a & PLM Adapter lightweight fine-tuning \\
          PLM\_a\_p & PLM Adapter lightweight Prompt-tuning \\
          \bottomrule
        \end{tabular}%
      }
      \caption{Abbreviations and Meanings}\label{tab:abb}%
    \end{table}%
    \subsection{Experimental Setup}
    This experiment was conducted on three machines with identical configurations. The relevant specifications and environment details can be found in Table \ref{tab:conf}.
    \begin{table}[ht!]
      \centering
      \resizebox{0.8\linewidth}{!}{
        \begin{tabular}{cc}
          \toprule
          Configuration Name & Parameters \\
          \midrule
          CPU & Intel Core i9-10900K CPU @ 3.70GHz \\
          GPU & NVIDIA Quadro RTX 4000 8GB \\
          Memory & 32GB \\
          Python Version & Python 3.8.16 \\
          PyTorch Version & 1.8.1+cu101 \\
          Operating System & Ubuntu \\
          \bottomrule
        \end{tabular}%
        }
      \caption{Hardware and Software Configurations}\label{tab:conf}%
    \end{table}%
    Due to insufficient GPU memory, experiments exploring CLP and CLW were not conducted in this study.
    \subsection{Hyperparameter Configuration}
    For the TNCC-t classification task, considering the relatively low character token count of each title in the dataset and the correspondingly small number of characters in the designed templates, we set the maximum length (Max\_len) to 108. During the CLA and CLAP tests, we chose a batch size (Batch\_size) of 4, while for other cases, the batch size was set to 16. It is worth noting that the learning rate (Learning\_rate) hyperparameter varies significantly in different scenarios, as shown in Table \ref{tab:lr}.
    \begin{table}[ht!]
      \centering
      \begin{tabular}{cc}
        \toprule
        Scenarios & Learning\_rate \\
        \midrule
        CSW, CBW, TW, TBW & $5 \times 10^{-6}$ \\
        CSP, CBP, TP, TBP & $6 \times 10^{-6}$ \\
        CSA, CBA, CLA & $1 \times 10^{-4}$ \\
        CSAP, CBAP, CLAP & $1.5 \times 10^{-4}$ \\
        TA, TBA & $3 \times 10^{-4}$ \\
        TAP, TBAP & $5 \times 10^{-4}$ \\
        \bottomrule
      \end{tabular}
      \caption{Learning Rates for Different  Scenarios}\label{tab:lr}
    \end{table}
    \subsection{Prompt-tuing Configuration}
\begin{figure}[ht!]
          \centering
          \includegraphics[width=1\linewidth]{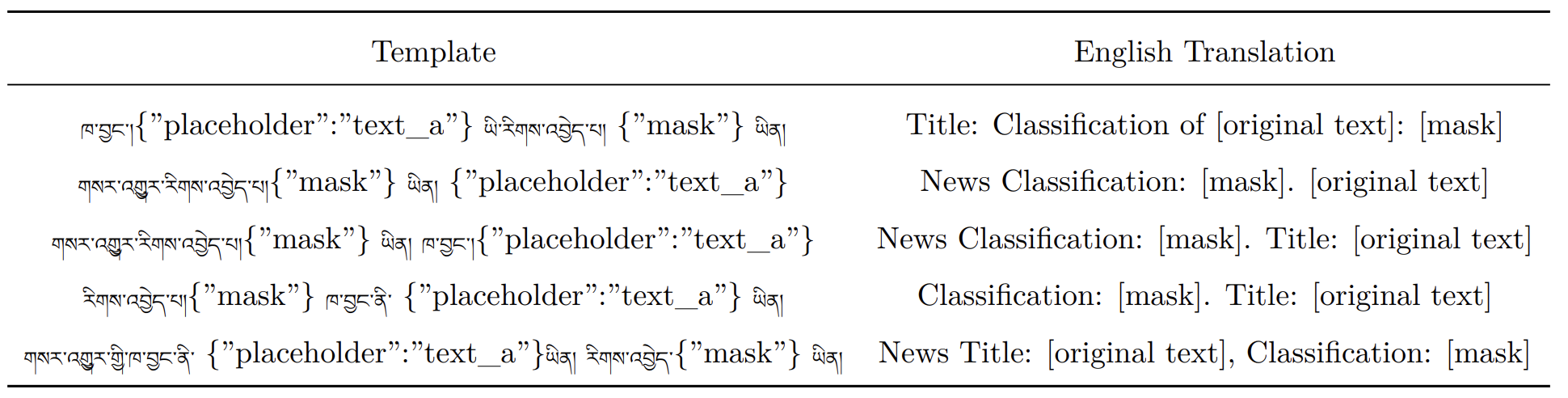} 
          \caption{\label{fig:temp} Templates and Translations} 
    \end{figure}

    We conducted Prompt-tuning experiments based on hard templates and validated the following effective templates. Please refer to Figure \ref{fig:temp} for details.
    
\begin{figure}[ht!]
          \centering
          \includegraphics[width=.8\linewidth]{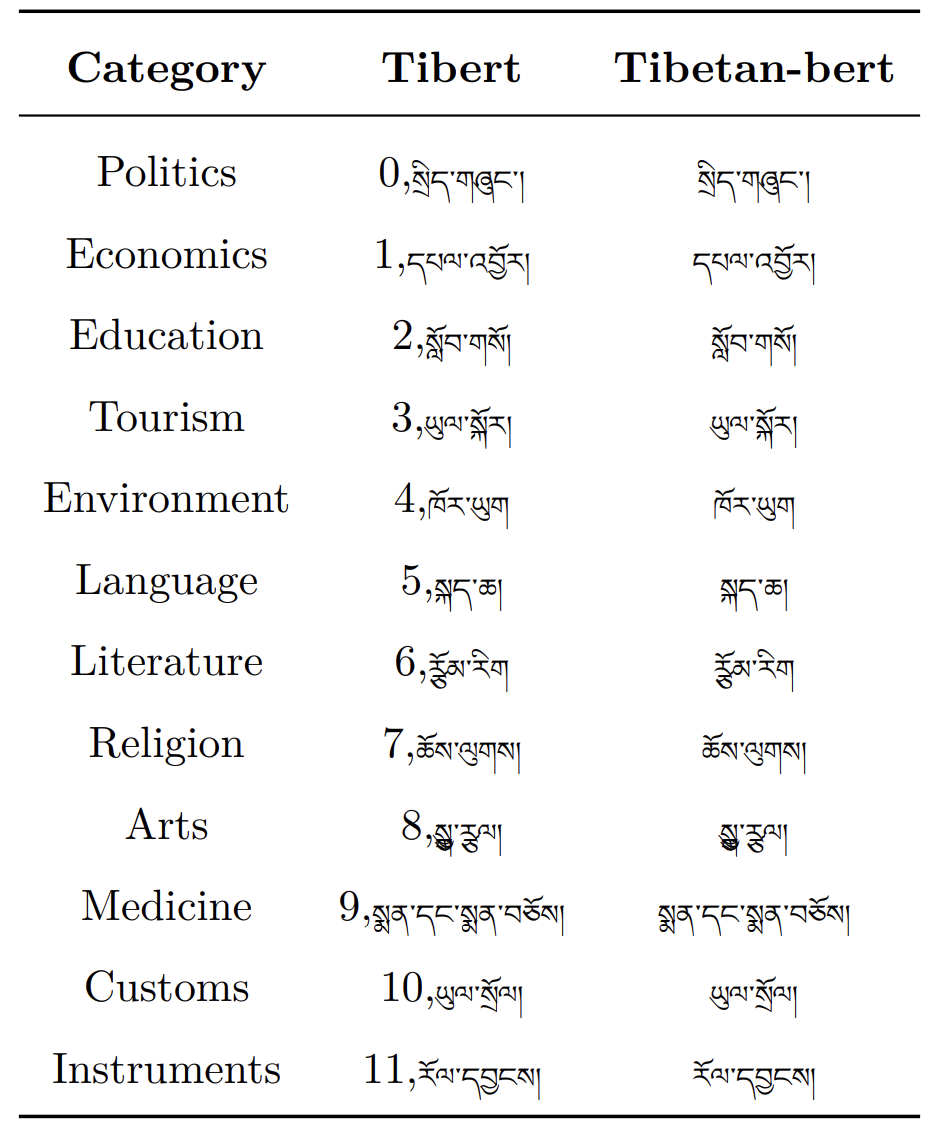} 
          \caption{\label{fig:vb} Verbalizer Configuration in the Experiment} 
    \end{figure}

    We chose the manual\_verbalizer approach. For CINO, as it is a multi-language pre-trained model capable of mapping English vocabulary, we selected English expressions corresponding to the original dataset labels as the verbalizer. For Tibert and Tibetan-bert, both dedicated Tibetan monolingual PLMs, we set the verbalizer to be in Tibetan. The verbalizer for Tibert was configured using a dictionary list format. The obtained Tibetan templates and verbalizers were translated from Tibetan to Chinese using the Bing platform.Please refer to Figure \ref{fig:vb} for the Verbalizer configuration in the experimental process.
    \subsection{Training Process}
    The training process varies slightly for each scenario, with the main differences being in the complexity of the preparation process before entering the training loop. We present the most complex TBAP training process (which involves injecting verbalizer-specific tokens) for demonstration. The entire process can be referred to in Algorithm \ref{algo:tbap}.
    \begin{algorithm}[ht!]
      \caption{TBAP Training Process}\label{algo:tbap}
      \begin{algorithmic}[1]
        \Require Training epochs $E$, training dataset $D$, validation dataset $V$
        \State Preprocess symbols in datasets $D$ and $V$
        \State Load PLM, tokenizer, model configuration, and WrapperClass
        \State Create twelve new Tibetan tokens corresponding to verbalizers and add them to the tokenizer's vocabulary
        \State Update token embeddings dimension of PLM
        \State Load local templates and verbalizers
        \State Create a train loader for $D$
        \State Inject adapters into PLM
        \For{$\text{epoch}$ from $1$ to $E$}
          \For{$\text{inputs}$ in $\text{train\_loader}$}
            \State Perform forward pass of inputs through PLM to obtain logits
            \State Compute cross-entropy loss for logits and their ground truth
            \State Perform backward pass of loss and update gradients
            \State Update parameters using Adam optimizer
          \EndFor
          \State Print loss, calculate and print $\text{val\_acc}$ and $\text{val\_f1}$ for $V$
        \EndFor
      \end{algorithmic}
    \end{algorithm}
    \subsection{Experimental Results}
    We conducted benchmark experiments with CSW, CBW, CLW, TW, and TBW as baseline scenarios. Except for CLW, we conducted tests at least five times for each scenario, except for CLP. The performance results for each scenario are presented in Table \ref{tab:com} .
    \begin{table}[ht!]
      \centering
      \resizebox{1\textwidth}{!}{%
      \begin{tabular}{ccccccc}
        \toprule
        Situation & Training Parameters & Vev\_acc & Vev\_macro\_f1 & Test\_acc & Test\_macro\_f1 & Training Parameters Ratio \\
        \midrule
        \multicolumn{7}{c}{CINO-small-v2}\\
        \toprule
        CSW & 147737868 & 0.67025 & 0.63686 & 0.69181 & 0.65737 & 1 \\
        CSP & 147865535 & \textbf{0.67133} {\scriptsize(+0.0054)} & \textbf{0.64198} {\scriptsize(+0.0067)} & \textbf{0.69827} {\scriptsize(+0.0043)} & 0.65840 {\scriptsize(+0.0141)} & 1.00086 \\
        CSA & 258048 & 0.66810 {\scriptsize(+0.0075)} & 0.61695 {\scriptsize(+0.0067)} & 0.67133 {\scriptsize(+0.0204)} & 0.63814 {\scriptsize(+0.01943)} & 0.174666\% \\
        CSAP & 258048 & 0.66918 {\scriptsize(+0.0010)} & 0.62477 {\scriptsize(+0.0196)} & 0.68426 {\scriptsize(+0.0119)} &\textbf{0.65917} {\scriptsize(+0.0063)} & 0.174666\% \\
        \toprule
        \multicolumn{7}{c}{CINO-base-v2}\\
        \toprule
        CBW & 190523148 & 0.68023 & 0.64022 & 0.68857 & 0.65310 & 1 \\
        CBP & 190650815 & \textbf{0.68750} {\scriptsize(+0.0032)} & \textbf{0.65806} {\scriptsize(+0.0017)} & \textbf{0.69612} {\scriptsize(+0.0065)} & \textbf{0.65955} {\scriptsize(+0.0079)} & 1.00067 \\
        CBA & 516096 & 0.65948 {\scriptsize(+0.0010)} & 0.59932 {\scriptsize(+0.0058)} & 0.67241 {\scriptsize(+0.0086)} & 0.62086 {\scriptsize(+0.0213)} & 0.270884\% \\
        CBAP & 516096 & 0.67025 {\scriptsize(+0.0107)} & 0.62589 {\scriptsize(+0.0200)} & 0.68211 {\scriptsize(+0.0107)} & 0.65416 {\scriptsize(+0.0067)} & 0.270884\% \\
        \toprule
        \multicolumn{7}{c}{CINO-large-v2}\\
        \toprule
        CLW & 443884556 & - & - & - & - & 1 \\
        CLP & 444009663 & - & - & - & - & 1.00028 \\
        CLA & 1376256 & \textbf{0.69827} {\scriptsize(+0.0064)} & \textbf{0.65242} {\scriptsize(+0.0135)} & 0.69073 {\scriptsize(+0.0021)} & 0.66243 {\scriptsize(+0.0053)} & 0.310048\% \\
        CLAP & 1376256 & 0.67887 {\scriptsize(+0.0140)} & 0.64039 {\scriptsize(+0.0094)} & \textbf{0.69504} {\scriptsize(+0.0032)} & \textbf{0.66635} {\scriptsize(+0.0104)} & 0.310048\% \\
        \toprule
        \multicolumn{7}{c}{Tibert}\\
        \toprule
        TW & 109610508 & 0.53448 & 0.47624 & 0.56573 & 0.52858 & 1 \\
        TP & 109632821 & \textbf{0.55711} {\scriptsize(+0.0129)} & \textbf{0.51984} {\scriptsize(+0.0046)} & \textbf{0.57327} {\scriptsize(+0.0021)} & \textbf{0.54577} {\scriptsize(+0.0090)} & 1.0002 \\
        TA & 516096 & 0.55387 {\scriptsize(+0.0409)} & 0.49573 {\scriptsize(+0.0649)} & 0.55711 {\scriptsize(+0.0075)} & 0.53333 {\scriptsize(+0.0066)} & 0.470845\% \\
        TAP & 516096 & 0.55603 {\scriptsize(+0.0150)} & 0.51313 {\scriptsize(+0.0132)} & 0.56788 {\scriptsize(+0.0194)} & 0.53690 {\scriptsize(+0.0075)} & 0.470845\% \\
        \toprule
        \multicolumn{7}{c}{Tibetan-bert}\\
        \toprule
        TBW & 11347724 & 0.63793 & 0.57721 & 0.65086 & 0.61974 & 1 \\
        TBP & 11372299 & \textbf{0.64655} {\scriptsize(+0.0006)} & \textbf{0.60635} {\scriptsize(+0.0034)} & \textbf{0.65840} {\scriptsize(+0.0075)} & \textbf{0.62614} {\scriptsize(+0.0117)} & 1.002165 \\
        TBA & 516096 & 0.58081 {\scriptsize(+0.0118)} & 0.51474 {\scriptsize(+0.0117)} & 0.59590 {\scriptsize(+0.0194)} & 0.55954 {\scriptsize(+0.0169)} & 0.463499\% \\
        TBAP & 516096 & 0.63469 {\scriptsize(+0.0096)} & 0.58375 {\scriptsize(+0.0252)} & 0.64008 {\scriptsize(+0.0150)} & 0.60228 {\scriptsize(+0.0200)} & 0.463499\% \\
        \bottomrule
      \end{tabular}}
      \caption{Comparison Results}\label{tab:com}
    \end{table}
    From Table \ref{tab:com}, the performance values for PLM\_a, PLM\_p, and PLM\_a\_p are composed of two parts: the average performance during experiments (represented in larger font) and the difference from the optimal performance (indicated by a plus sign and smaller font). Based on the information in Table \ref{tab:com}, the following conclusions can be drawn:
    For each Tibetan PLM studied in this experiment, PLM\_p generally exhibits the best overall performance when considering a slight increase in training parameters and the incorporation of a small amount of template text. PLM\_p shows a 1.6\% to 2.4\% improvement in Test\_macro\_f1 compared to PLM\_wo.
    The performance metrics for PLM\_a are usually lower than those of PLM\_wo. However, considering the significant reduction in the number of training parameters, PLM\_a's performance metrics remain competitive. For example, the average performance of CSA is at most around 2\% lower than CSW, and the highest performance only has a 1.8\% difference in Dev\_macro\_f1. However, its training parameters are only 0.17\% of CSW's.
    TA's Test\_macro\_f1 performance is even higher than TW. Considering the combination of both Prompt-tuning and Adapter techniques, the results for PLM\_a\_p match our expectations. PLM\_a\_p achieves remarkable results by significantly reducing the number of training parameters while substantially improving performance. From Table \ref{tab:com}, it can be observed that PLM\_a\_p generally achieves results that meet or even exceed those of PLM\_wo. The average Test\_macro\_f1 values for CSAP, CBAP, and TAP are all higher than those for CSW, CBW, and TW, respectively.
    \begin{figure}[ht!]
          \centering
          \includegraphics[width=.8\linewidth]{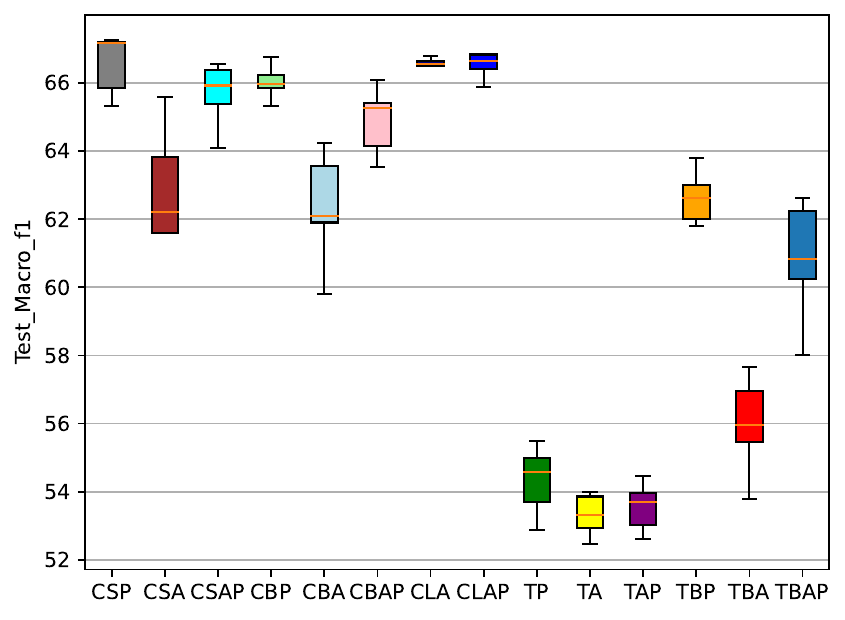} 
          \caption{\label{fig:effect} The average performance of efficient fine-tuning} 
    \end{figure}
    To visually illustrate the performance enhancement effects of the three efficient fine-tuning techniques (PLM\_p, PLM\_a, and PLM\_a\_p) on PLMs and their respective performance relationships, please refer to Figure \ref{fig:effect}. Each small box in the figure \ref{fig:effect} corresponds to the average performance of the five experiment data samples for each of the three scenarios in this experiment.
    
    \section{Conclusion}
    In the context of Tibetan PLMs, this study investigated three efficient fine-tuning methods: Prompt-tuning, Adapter, and the combination of Adapter with Prompt-tuning. Through extensive experimentation, we have demonstrated that Prompt-tuning has a positive impact on enhancing PLM performance, while Adapter significantly reduces the number of training parameters, achieving lightweight fine-tuning. By combining these two fine-tuning techniques, we achieved a balanced effect that incorporates the advantages of both.
    
    \section{Acknowledgments}
    This research work was supported by the following grants: the National Key R\&D Program of China(No. 2022ZD0116100), the National Natural Science Foundation of China (No. 62162057),and the Everest Discipline Construction Project of Tibet University under Project (No. zf22002001).
    \bibliography{PEFTT}
    \bibliographystyle{elsarticle-num}
\end{document}